\documentclass[letterpaper]{article} 
\usepackage{aaai25}  
\usepackage{times}  
\usepackage{helvet}  
\usepackage{courier}  
\usepackage[hyphens]{url}  
\usepackage{graphicx} 
\usepackage{natbib}  
\usepackage{booktabs} 
\usepackage{caption} 
\usepackage{algorithm}
\usepackage{algorithmic}

\usepackage{newfloat}
\usepackage{listings}
\DeclareCaptionStyle{ruled}{labelfont=normalfont,labelsep=colon,strut=off} 
\floatstyle{ruled}
\newfloat{listing}{tb}{lst}{}
\floatname{listing}{Listing}
\title{Auto-FEDUS: Autoregressive Generative Modeling of Doppler Ultrasound Signals from Fetal Electrocardiograms}
\author{
    Alireza Rafiei\textsuperscript{\rm 1},
    Gari D. Clifford\textsuperscript{\rm 1,2},
    Nasim Katebi\textsuperscript{\rm 1}
}
\affiliations{
    \textsuperscript{\rm 1}Emory University, USA\\
    \textsuperscript{\rm 2}Georgia Institute of Technology, USA\\
}

\usepackage{bibentry}

\begin{document}

\maketitle

\begin{abstract}
Fetal health monitoring through one-dimensional Doppler ultrasound (DUS) signals offers a cost-effective and accessible approach that is increasingly gaining interest. Despite its potential, the development of machine learning based techniques to assess the health condition of mothers and fetuses using DUS signals remains limited. This scarcity is primarily due to the lack of extensive DUS datasets with a reliable reference for interpretation and data imbalance across different gestational ages. 
In response, we introduce a novel \textbf{auto}regressive generative model designed to map fetal electrocardiogram (\textbf{FE}CG) signals to corresponding \textbf{DUS} waveforms (\textit{Auto-FEDUS}). By leveraging a neural temporal network based on dilated causal convolutions that operate directly on the waveform level, the model effectively captures both short and long-range dependencies within the signals, preserving the integrity of generated data. 
Cross-subject experiments demonstrate that Auto-FEDUS outperforms conventional generative architectures across both time and frequency domain evaluations, producing DUS signals that closely resemble the morphology of their real counterparts. 
The realism of these synthesized signals was further gauged using a quality assessment model, which classified all as good quality, and a heart rate estimation model, which produced comparable results for generated and real data, with a Bland-Altman limit of 4.5 beats per minute. This advancement offers a promising solution for mitigating limited data availability and enhancing the training of DUS-based fetal models, making them more effective and generalizable.
\end{abstract}

%

\section{Introduction}
Despite significant global medical advancements that have lowered mortality rates in many countries, childbirth still presents considerable risks to both mothers and children. Each year, approximately 2 million stillbirths and 2.3 million neonatal deaths occur, with low- and middle-income countries (LMICs) accounting for about 98\% of these perinatal losses \cite{UNIGME2020, UNIGME2021}. 
Various pathophysiological factors, such as congenital heart defects (CHD), fetal growth restriction (FGR), and conditions that lead to fetal hypoxia, contribute to fetal and maternal health risks. The early detection of these conditions through fetal monitoring plays a crucial role in preventing adverse outcomes. 
Nonetheless, inadequate infrastructure and a shortage of trained healthcare professionals hinder the identification of health risks and implementation of potential interventions, especially in LMICs and rural regions.

While ultrasound imaging is a favored method for monitoring fetal health in well-resourced regions, cost-effective alternatives for perinatal screening using one-dimensional Doppler ultrasound (DUS) and fetal electrocardiogram (FECG) signals are emerging in other areas \cite{gibb_fetal_2023, katebi_hierarchical_2023}. These non-invasive, low-cost techniques, especially when augmented with machine learning and edge computing, have the potential to reduce reliance on expensive infrastructure and highly trained medical professionals in low-resource settings, lower monitoring costs, and significantly enhance accessibility. DUS signals are adept at capturing dynamic blood flow information from the fetus’s cardiac activity. They provide a detailed view of the mechanical aspects of heart function, the opening and closing of the valves, and insights into the cardiovascular status of the fetus. 
Such information is invaluable for detecting fetal hypoxia, CHD, FGR, and evaluating placental circulation, whether through automated detection or manual evaluation, tasks that are difficult to obtain directly from other methods \cite{marzbanrad_cardiotocography_2018, stroux_dopplerbased_2017}. 
Additionally, hybrid approaches that combine simultaneous DUS and FECG signals for various fetal health monitoring yielded promising results \cite{10.3389/fphys.2017.00313,yamamoto_ecg_2020}.

Despite the abundance of diverse datasets for ECG signals, there is a paucity of DUS data with clinical diagnosis, particularly during earlier stages of pregnancy \cite{Sulas2021}. This limitation is further compounded by challenges such as the duration of good quality signal acquisition and the lack of accurate event labels. While several generative approaches exist for producing synthetic FECG signals under various conditions \cite{sameni2007multichannel}, the literature on generating synthetic DUS data is minimal \cite{Queyam2017,Pennati1997,Garcia-Canadilla2014}.
This leaves a significant gap in the development of machine learning solutions for DUS-based fetal health analysis. 

To address this challenge, we introduce \textit{Auto-FEDUS}, a novel end-to-end \textbf{auto}regressive network designed to take \textbf{FE}CG as input and generate corresponding high-fidelity \textbf{DUS} waveforms. We observe that conventional generative adversarial networks (GANs) face challenges in capturing the nuances of bio-signal extrapolation and achieving a balance between the generator and discriminator. They tend to overfit on the discriminator rapidly, necessitating the use of a simpler discriminator network compared to the generator, particularly when working with insufficient data and varying frequencies. They also struggle to identify regions of interest, which prevents them from learning both global structures and fine-grained local details. Although various architectural and training strategies have been proposed to mitigate these issues, they do not consistently deliver the desired results across all use cases. \cite{arjovsky_wasserstein_2017,karras_progressive_2018}. Additionally, developing parametric models for this extrapolation is complex and lacks generalizability. To overcome these issues, Auto-FEDUS leverages an autoregressive approach with dilated causal convolutions, which allow the model to efficiently capture both short and long-range temporal dependencies by expanding the receptive field without increasing computational complexity. This enables the model to simultaneously learn the intricate patterns of DUS in both time and frequency domains and facilitates continuous improvement with new data. 

The primary motivation behind this study is to leverage available FECG datasets and methods to synthesize FECG signals to generate realistic heartbeat DUS signals, thereby augmenting the DUS datasets. The idea is to learn the cross-modality signal-to-signal extrapolation by capturing the inherent correlation between electrical and mechanical cardiac activities. This can facilitate the development of machine learning models and enhance the generalization of parametric and statistical methods that employ DUS or hybrid approaches combining DUS and FECG for a wide range of fetal health monitoring applications. Below is a summary of our main contributions:

\begin{itemize}
\item We present a generative model for high-fidelity FECG-to-DUS signal mapping. To the best of our knowledge, this is the first generative model for this cross-modal signal extrapolation, for DUS signal generation, and for different frequency handling in the bio-signal domain.

\item We tackle the challenge of extrapolating low- to high-frequency signals by introducing an autoregressive model that works directly on one-dimensional time-series data. We also develop a wide range of different generative models and GANs with a game-theoretic training approach
for performance comparison.

\item In addition to conducting a range of subject-independent analyses to assess the quality of generated signals across different domains, we evaluate these DUS waveforms using established, practical, and robust fetal health monitoring models.
\end{itemize}

\section{Related Work}
The research landscape appears under-explored in both the generation of synthetic DUS signals using generative methods and the specific challenge of extrapolating bio-signals to DUS; to our knowledge, there has been no prior published work in these areas. Nonetheless, several studies have focused on mapping one physiological signal to another, primarily involving low-frequency signals with the same handling sampling frequency. 
Earlier investigations into the inherent patterns between PPG and ECG focused on estimating different ECG parameters from PPG signals using machine learning-based methods \cite{banerjee2014photoecg}, progressing to reconstructing the ECG signal from PPG using techniques such as regression \cite{zhu2021learning} and dictionary learning \cite{tian2020cross}. Such methods experience a decline in performance when evaluated in a cross-corpus and subject-independent manner. Later, \cite{sarkar_cardiogan_2021} presented an adversarial framework, CardioGAN, that utilized an attention-based U-Net generator and a dual time-frequency discriminator to maintain the integrity of data generation. \cite{vo_p2e-wgan_2021} and \cite{kong2024f} proposed P2E-WGAN and f-GAN, respectively, both featuring a U-Net-based generator and a convolutional neural network (CNN)-based discriminator. \cite{abdelgaber_subject-independent_2023} developed a lightweight autoencoder to tackle the computational capabilities of wearable devices, and \cite{chen2023deep} proposed a CycleGAN-based model for beat-to-beat waveform signal conversion. Recently, \cite{vo2023ppg} introduced an attention-based deep state-space model to incorporate prior knowledge of signal structures, and \cite{shome2024region} proposed a novel diffusion model for the extrapolation. 
In addition to mapping PPG to ECG signals, other physiological signal conversions have been explored. \cite{harfiya2021continuous} and \cite{ibtehaz2022ppg2abp} developed deep learning-based methods to map PPG signals to arterial blood pressure (ABP) waveforms. 
Furthermore, \cite{yu_improving_2023} proposed an attention-based Wasserstein GAN (WGAN) to generate low-frequency Doppler radar signals for non-contact heart rate monitoring from ECGs to address the issue of data imbalance. Here, however, we focus on synthesizing DUS from FECG signals. Unlike the majority of the mentioned methods, which used GAN-based approaches, worked with low-frequency signals, and matched the input and output sampling frequencies, our autoregressive model features low computational complexity and extrapolates signals from a lower to a higher frequency.

\section{Methodology}
\subsection{Study's Workflow}

\begin{figure}[h]
  \centering
  \includegraphics[width=0.475\textwidth]{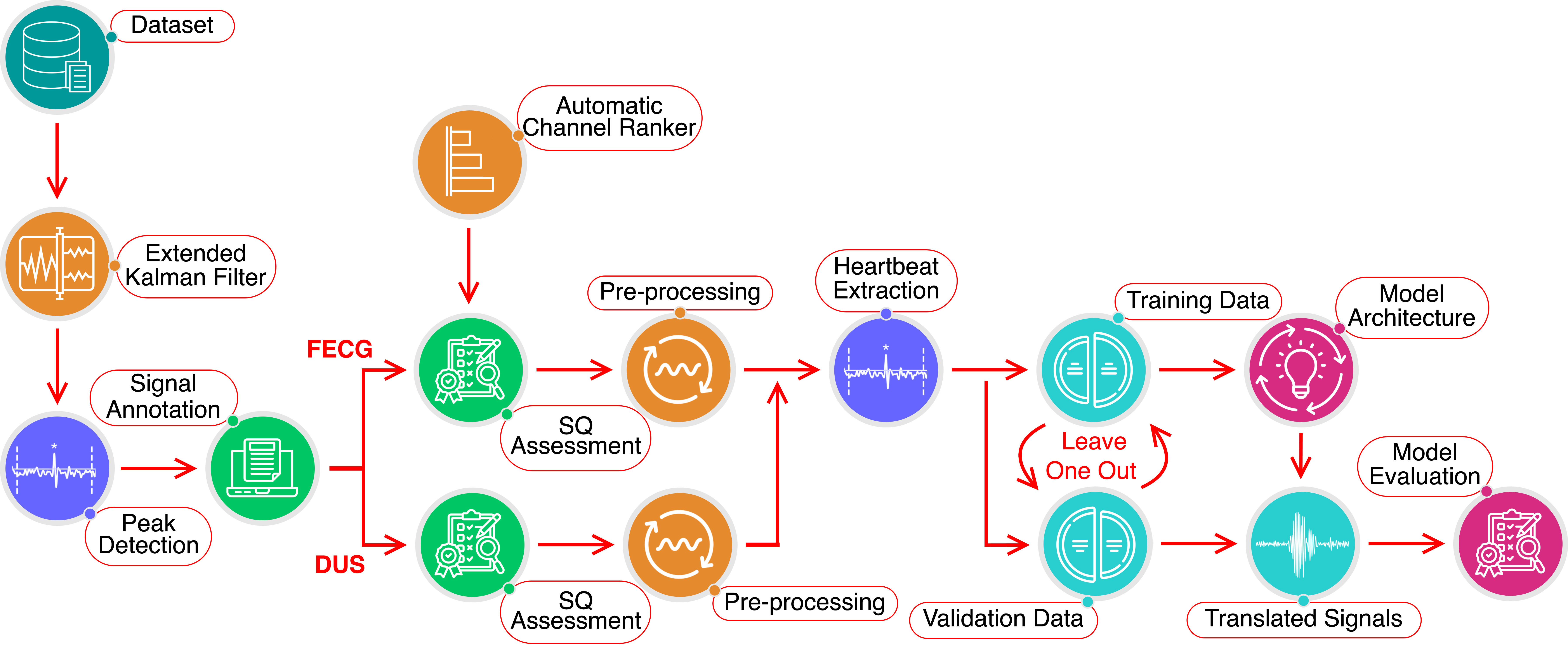}
  \caption{Auto-FEDUS development workflow from data collection to model development.}
  \label{fig:workflow}
\end{figure}

Figure \ref{fig:workflow} provides a general overview of the study's workflow. The model development process began with collecting a dataset comprising abdomen ECG signals and their corresponding simultaneous DUS data. The raw ECG signals were initially filtered to extract FECG. After filtering, FECG peaks were automatically identified and subsequently adjusted by human annotators, who also assessed the signal quality index (SQI) of all signals using a developed MATLAB graphical user interface (Appendix A). Based on the provided indices, only signals meeting the specified quality criteria were advanced to the next stage. Each signal modality then underwent additional processing and heartbeats were extracted using the identified peak locations. The resulting data was ultimately split into training and validation sets to develop and validate a model capable of reliably transforming FECG into DUS signals.

\subsection{Auto-FEDUS Framework}
\begin{figure}[h]
  \centering
  \includegraphics[width=0.5\textwidth]{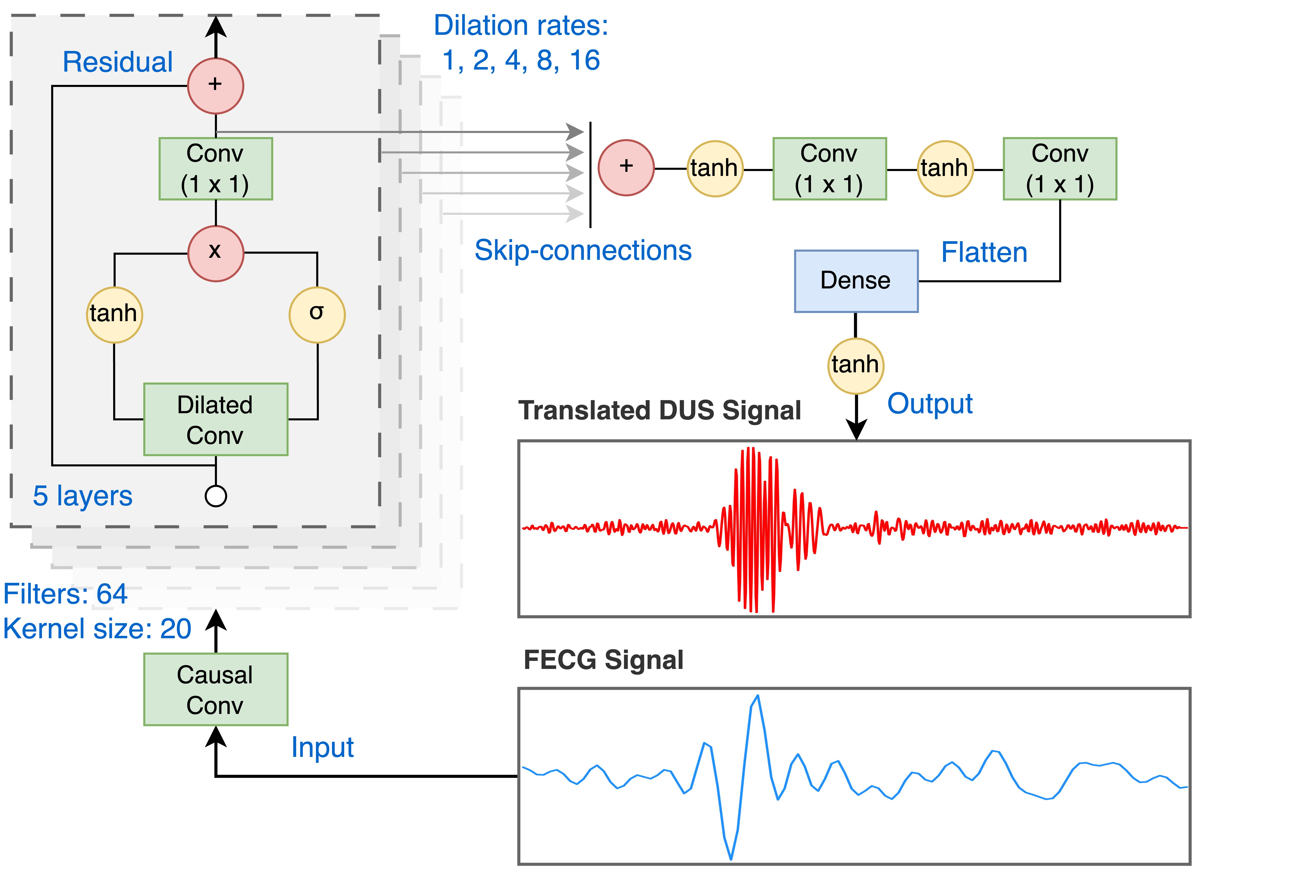}
  \caption{Auto-FEDUS's architecture for generative mapping of FECG-to-DUS signal.}
  \label{fig:Auto-FEDUS}
\end{figure}

Extrapolating low-frequency signals with relatively predictable and stable patterns into high-frequency signals, often characterized by abrupt variations, requires special considerations to capture their dynamics. 
Here, inspired by a text-to-speech model called WaveNet \cite{oord2016wavenet}, we introduced Auto-FEDUS, an end-to-end deep generative model for physiological cardiac cycle signal-to-signal mapping. The architecture of the model, which is shown in Figure \ref{fig:Auto-FEDUS}, begins with a causal convolution layer that processes the input FECG signal. This layer ensures that the output at any given time step solely depends on the current and previous inputs to preserve data sequentially. The causal convolution uses 64 filters with a kernel size of 20, extracting initial features from the input signal. After this initial convolution, the model employs a series of five residual blocks, each consisting of dilated convolutions with the same number of filters and kernel size. These blocks are designed to capture long-range dependencies within the signal while maintaining computational efficiency. The dilations have rates of 1, 2, 4, 8, and 16, progressively increasing to expand the receptive field without the need for stacking layers. 

Within each residual block, the output from the dilated convolution is passed through non-linear activation functions (hyperbolic tangent and sigmoid), and then combined multiplicatively. After passing the results through another convolutional layer, the resulting features are added back to the block's input via a residual connection, which helps mitigate the vanishing gradient problem during training. These features of each residual block are also combined through skip connections, which are summed and passed through additional convolution layers. These layers further refine the features before they are flattened and processed by the final dense layer. The output layer of the model uses a hyperbolic tangent activation function to generate the extrapolated DUS signal.

\subsection{Related Generative Models}
Since no previous study has focused on the generative mapping of FECG to DUS signals, we developed several different generative approaches to examine their performance and compare them with Auto-FEDUS. Our baseline model was an autoencoder, whose decoder consisted of three convolutional layers, each with a kernel size of eight and a stride of one. After the first convolutional layer, we applied max pooling, and batch normalization followed each subsequent layer. The encoder featured three convolutional blocks, each containing a convolutional layer with the same kernel and stride configurations, along with upsampling and batch normalization. Next, we designed a GAN model based on long short-term memory (LSTM). In this model, both the generator and discriminator comprised three LSTM layers. Another model we explored was a deep convolutional GAN (DCGAN), which includes three transposed convolutional layers with a kernel size of eight and a stride of two, each layer followed by batch normalization. The discriminator in this model employed two convolutional layers, both with a kernel size of eight and a stride of one, with batch normalization applied after each layer. Lastly, we developed a WGAN with gradient penalty (WGAN-GP) model in which the generator included three transposed convolutional layers with a kernel size of twelve and a stride of two, each followed by batch normalization \cite{arjovsky_wasserstein_2017}. The discriminator also consisted of three convolutional layers with the same kernel and stride sizes, accompanied by batch normalization; the first two layers also incorporated max pooling with a size of two. 

\section{Experiments}

\subsection{Dataset}
The dataset used in this study was collected at the Leipzig University Hospital in Germany \cite{valderrama_open_2019}. It comprises one-dimensional DUS signals and simultaneously recorded seven-channel indirect abdominal ECG with a maternal ECG reference from five volunteers in the 20th to 27th weeks of their pregnancies. The DUS signals were captured using a hand-held device 
operating at an ultrasound frequency of 3.3 MHz and a digital sampling frequency of 44.1 kHz. The abdominal ECG signals were recorded using a 16-bit commercial analog-to-digital converter 
and stored at a sampling frequency of 1000 Hz with applying spectral filtering in the hardware
with a cutoff frequency of 50 Hz. The ethical approval for the study was granted by the ethics committee of the Leipzig University Hospital, and written informed consent was secured from each participant.

We manually annotated the dataset to assess the quality of FECG and DUS signals and to correct the automatic FECG peak detections. Prior to annotation, DUS signals were downsampled to 4k Hz using an anti-aliasing filter, FECG was extracted from the seven-channel abdomen ECG using an extended Kalman filter, and the location of FECG peaks for each channel was estimated with the FECGSYN toolbox \cite{andreotti_open-source_2016}. During the annotation process, two annotators adjusted the automatically detected FECG peak locations and assessed the signal quality of each 3.75\,s segment of the DUS and each FECG channel
As a result, each DUS and FECG channel segment received an SQI ranging from 1 to 5 by each annotator independently. SQIs from 1 to 5 for DUS represent good, poor, interference, talking, and silent segments and for FECG indicate clean, irregular, noisy, heavy noise, and unsure segments, respectively. 



\subsection{Data Preparation}
Considering the characteristics of physiological time-series data, DUS and FECG signals often contain distortions due to both internal and external interferences, such as respiratory activities, patient movement, and environmental noise. In response, we leveraged the available SQI annotations for each 3.75\,s segment—a standard duration in cardiotocography that balances adequate data collection with signal stationarity \cite{valderrama_open_2019}—and processed both signals. Specifically, we only kept DUS signals with an SQI of 1, which is an indication of good quality. 
To maximize data retention, we adopted a fusion strategy combined with an automatic signal quality ranker to identify the most suitable FECG channel for the corresponding DUS. That is, we first calculated and ranked each FECG channel based on eight different automatic quality measures proposed by OSET \cite{sameni2012oset}. In the fusion process, if both annotators rated the quality of one or more FECG channels 1, we selected the channel with the highest automatic ranking among those rated as 1. When none of the channels received a quality score of 1 from one annotator, we summed the two SQIs for each channel and chose the channel with the lowest total score. If multiple channels shared the lowest total, we chose the one with the lowest individual SQIs; if no clear choice emerged, the highest automatic ranking was selected. We excluded signal pairs from our analysis if the combined SQI of both annotators exceeded 5 or if either annotator assigned an SQI of 5. As a result, the available FECG signal was not confined to any specific channel for a corresponding DUS, allowing the models to generate DUS signals independently of the FECG channel.

Afterward, we resampled the DUS signals to 2k Hz and the FECG data to 250 Hz for all subjects. We then applied a second-order Butterworth bandpass filter to the DUS waveforms with cut-off frequencies of 25 and 600 Hz \cite{valderrama_open_2019}. For the FECG signals, we used a band-pass finite impulse response (FIR) filter, setting the pass-band frequency at 3 Hz and the stop-band frequency at 45 Hz. Next, we performed segment-wise normalization and investigated possible lags between the simultaneously recorded signals. For this aim, we calculated the homomorphic envelopes for DUS and the Pan-Tompkins envelopes for FECG signals and then computed the cross-correlation between these envelope pairs. By identifying the highest cross-correlation values, we estimated and adjusted the lags between the pair of signals for each subject separately (Appendix A). Finally, we extracted every FECG and their corresponding DUS beats using the available peak placements of the 3.75\,s segments and applied min-max normalization to both signals to ensure all input data falls within a specific range. In the end, 3923 FECG and their corresponding DUS heartbeats were prepared for further experiments.

\subsection{Training Details}
The training and evaluation were conducted using a subject-independent scheme, where the models were trained on signals from four subjects and then tested on data from an unseen subject; this process was repeated five times. 
The Adam optimizer and mean squared error loss function were utilized to train Auto-FEDUS. 
For the GAN-based models, weights were initialized with a truncated normal distribution with a mean of 0 and variance of 0.02, and the WGAN-GP gradient penalty coefficient was set to 10. The generative models were implemented using TensorFlow 2.15.0 and Python 3.11.5. For training and testing the models, we utilized a computing system equipped with 64 GB of RAM, a single 14‑core CPU, and one NVidia Tesla P100 GPU. The processing time for mapping FECG to DUS signal was approximately 0.97 milli seconds per recording. 

\begin{table*}[h!]
\centering
\caption{Comparison of various evaluation indicators for the developed generative models using leave-one-out cross-validation.}
\label{table:per}
\fontsize{9}{11}\selectfont
\begin{tabular}{lcccccccc}
\toprule
\textbf{Model} & \textbf{RMSE} & \textbf{MAE} & \textbf{KLD} & \textbf{SE} & \textbf{PSDD} & \textbf{CD} & \textbf{SF} & \textbf{FD} \\ \midrule
Autoencoder & 1.66$\pm$1.16 & 0.83$\pm$0.51 & 2.20$\pm$1.74 & 2.29$\pm$0.16 & 13.94$\pm$16.14 & 241.58$\pm$58.73 & 0.02$\pm$0.03 & 11.45$\pm$2.52 \\ 

LSTMGAN & 0.98$\pm$0.04 & 0.96$\pm$0.04 & 5.46$\pm$0.22 & 3.47$\pm$0.31 & 0.37$\pm$0.14 & 342.34$\pm$79.44 & 0.01$\pm$0.00 & 20.57$\pm$1.54 \\ 

DCGAN & 0.23$\pm$0.04 & 0.16$\pm$0.03 & 0.70$\pm$0.88 & 0.99$\pm$0.67 & 0.39$\pm$0.01 & 104.30$\pm$30.54 & 0.06$\pm$0.02 & 5.14$\pm$0.63 \\ 

WGAN-GP & 0.22$\pm$0.01 & 0.15$\pm$0.01 & \textbf{0.03}$\pm$0.02 & 0.41$\pm$0.25 & 0.29$\pm$0.16 & 93.73$\pm$79.79 & 0.08$\pm$0.01 & 4.70$\pm$0.40 \\ 

\textbf{Auto-FEDUS} & \textbf{0.20}$\pm$0.01 & \textbf{0.14}$\pm$0.01 & 0.05$\pm$0.01 & \textbf{0.25}$\pm$0.27 & \textbf{0.27}$\pm$0.14 & \textbf{84.52}$\pm$43.37 & \textbf{0.00}$\pm$0.00 & \textbf{4.63}$\pm$0.27 \\ 
\bottomrule
\end{tabular}
\label{tab:performance}
\end{table*}

\subsection{Evaluation Indicators}
To quantitatively evaluate and comprehensively compare the developed models, we used eight different metrics to assess each pair of real DUS and its synthetic counterpart in both the time and frequency domains.

\textit{Root Mean Square Error (RMSE):}
Provides a measure of the magnitude of the error between generated and real signals. We calculated RMSE as $\sqrt{\frac{1}{n} \sum_{i=1}^{n} \left( x_i - \tilde{x}_i \right)^2}$, where $x_i$ and $\tilde{x}_i$ refer to the $i$\textsuperscript{th} sample of real and generated signals, respectively.

\textit{Mean Absolute Error (MAE):}
Quantifies the average magnitude of errors between paired generated and real signals without considering their direction. It's less sensitive to outliers than RMSE and was calculated by $\frac{1}{n} \sum_{i=1}^{n} \left| x_i - \tilde{x}_i \right|$.

\textit{Kullback-Leibler Divergence (KLD):}
A measure of how the probability distribution of the generated signals, $P_g(x)$, diverges from that of the real DUS, $P_r(x)$, calculated as $\sum P_r(x) \log \left( \frac{P_r(x)}{P_g(x)} \right)$.

\textit{Spectral Entropy (SE):}
An absolute difference in spectral entropy, which is a measure to quantify the complexity or predictability of a signal's power spectrum distribution, between the generated and real signals. SE was calculated as $- \sum p_j \log(p_j)$, where $p_j$ are the normalized powers at each $j^{th}$ frequency component.

\textit{PSD Difference (PSDD):}
Computes the Euclidean distance between the averaged power spectral densities (PSD) of the 
signals in the unit of millidecibels per Hz.

\textit{Centroid Difference (CD):}
Calculates the absolute difference in spectral centroids, which indicate the center of mass of the spectrum, between the generated and real signals.

\textit{Spectral Flatness (SF):}
Evaluates the flatness or tonality of a signal's power spectrum. It was computed as the ratio of the geometric mean to the arithmetic mean of the power spectrum values; a higher value indicates a more noise-like signal.

\textit{Fréchet Distance (FD):}
A measure used to quantify the similarity between two signals by considering the location and order of the points along the paths. We calculated FD based on the description in \cite{sarkar_cardiogan_2021}.

\begin{figure}[h!]
  \centering
\includegraphics[width=0.48\textwidth]{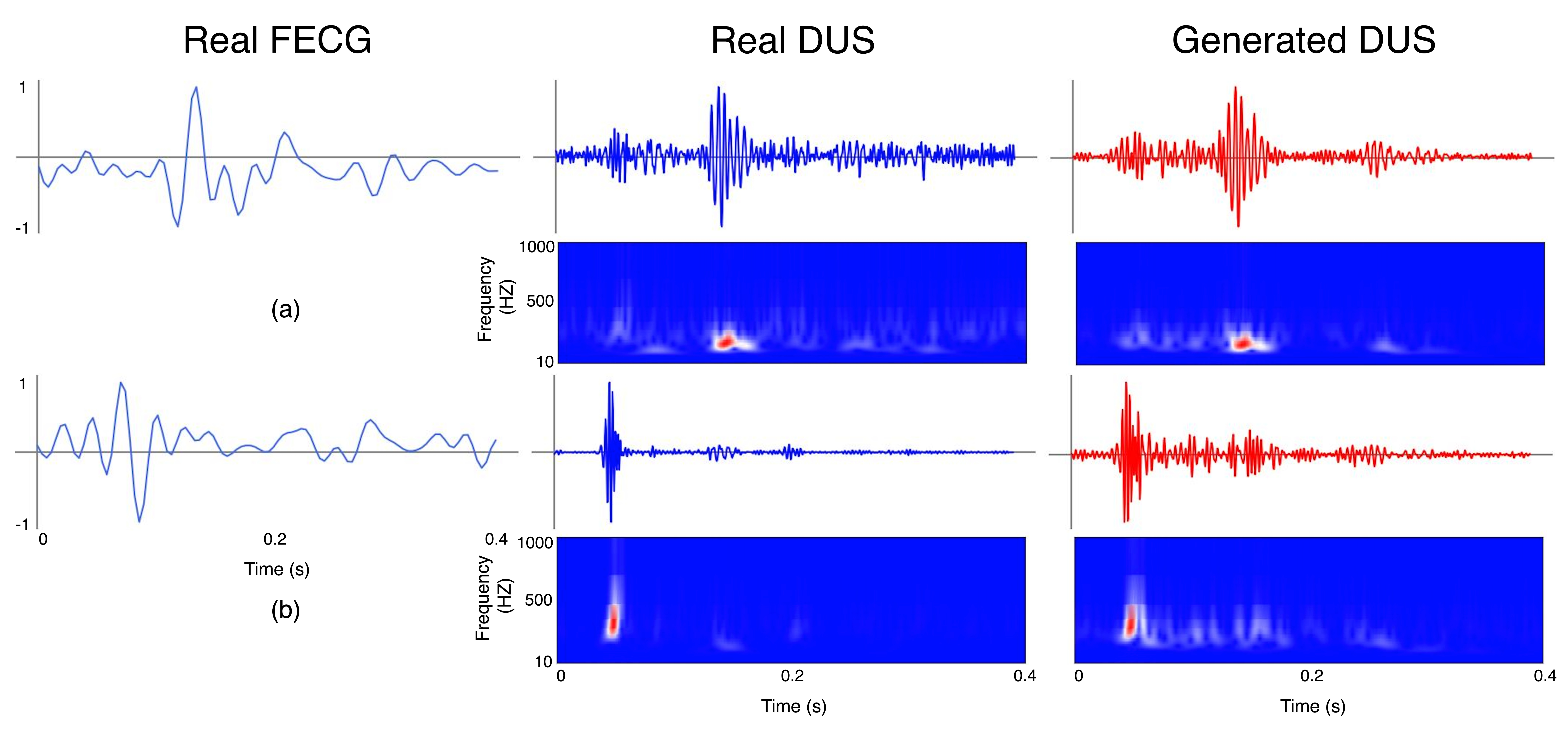}
  \caption{Comparison of real and generated signals for two random samples. Each panel presents a real FECG signal (left), its corresponding real DUS signal (middle), and the generated DUS signal (right), along with their respective scalograms.}
  \label{fig:samples}
\end{figure}

\begin{figure*}[h!]
  \centering
  \includegraphics[width=0.93\textwidth]{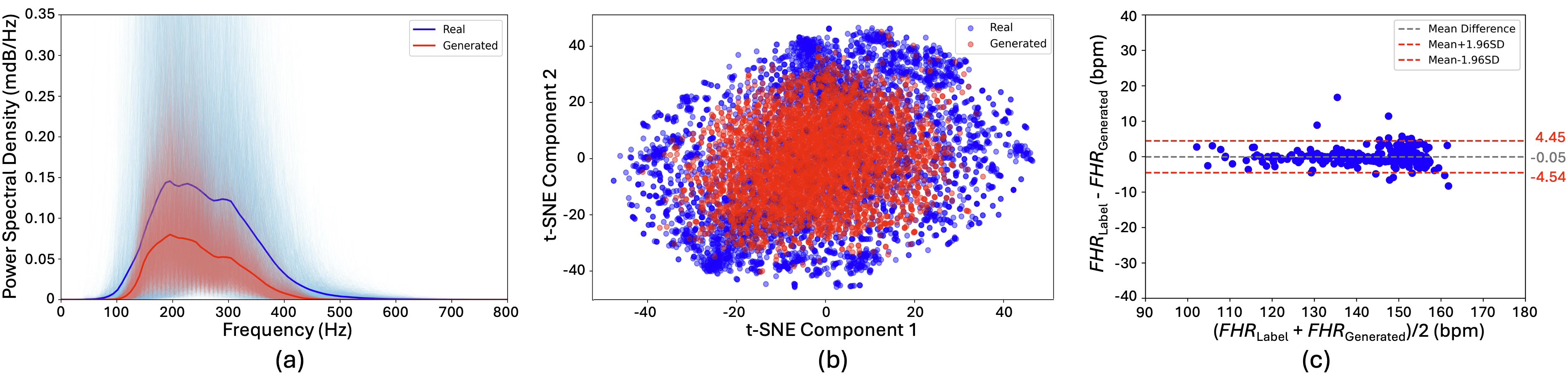}
  \caption{Real and generated data characteristics. (a) PSD analysis (b) t-SNE visualization (c) Bland-Altman plot. \textit{FHR}\textsubscript{Label} is derived from the simultaneously recorded FECG of each real DUS segment, and \textit{FHR}\textsubscript{generated} is estimated by the model.}
  \label{fig:PSD}
\end{figure*}

\section{Performance}
\subsection{Qualitative Evaluation}
Figure \ref{fig:samples} presents two sets of qualitative samples of the real and model's generated heartbeat DUS signals along with their corresponding FECG signals. To further analyze the characteristic frequency shifts in these signals over time, we have included their scalograms. The scalogram provides a two-dimensional representation of the signal, constructed as a function of time and frequency, based on the absolute values derived from the continuous wavelet transform. Although events are visible in the time domain, frequency domain analysis reveals the signal components at various frequencies, showing how the signal's energy is distributed across these frequencies. We can observe in Figure \ref{fig:samples} (a) and (b) that the model's generated DUS signals visually resemble the real DUS waveforms in terms of morphology and timing. The scalograms of the generated DUS signals display patterns similar to those of the real DUS scalograms, suggesting that the model effectively captures both the frequency content and its temporal variations. However, in some cases, 
the generated signals may not match the real DUS signals. This discrepancy primarily arises when the real FECG or DUS signals contain some levels of inherent noise, which complicates the model's ability for accurate extrapolation. Additionally, DUS signals are highly variable; even for relatively similar FECG signals, their structures can vary over time. A larger dataset from more subjects might mitigate this challenge. It should be noted that among the developed models, the Autoencoder and LSTMGAN models are unable to generate morphologically similar signals, whereas the other models can capture the general shape and characteristics during signal generation.

\subsection{Quantitative Evaluation and Analysis}
Table \ref{table:per} details the results of evaluation indicators used to assess the structural similarity between signals generated by the models and the original DUS signals. That is, we compared the performance of Auto-FEDUS with other commonly used generative models for mapping FECG to DUS signals across different domains. 
The reported results represent the mean and standard deviation of the metrics, based on leave-one-out cross-validation. As shown, the Auto-FEDUS model demonstrates superior performance across various metrics, including RMSE, MAE, SE, PSDD, CD, SF, and FD, whereas WGAN-GP excels in KLD.

Figure \ref{fig:PSD} (a) depicts the PSD representation of DUS heartbeat signals and their corresponding generated waveforms as a function of frequency using a subject-independent scheme. The PSD measures the power distribution of a signal across different frequency components. The pale colors represent individual signals, while the bold lines indicate their mean. The real data exhibits higher PSD values in the range of approximately 150 to 400 Hz, with a peak of around 200 Hz. The generated data shows a similar pattern and trend with nearly the same peak location but generally lower PSD values, indicating that the generated data approximates the real data's frequency characteristics with lower power. Figure \ref{fig:PSD} (b) provides a t-distributed stochastic neighbor embedding (t-SNE) visualization of the real and generated signals to compare their distributional characteristic. This method reduces the complex, high-dimensional feature space of these signals into a more interpretable two-dimensional plot to facilitate a visual assessment of similarity and variability between the signals. Each dot in this plot represents an individual signal. The plot shows a broad distribution where both real and generated data are spread across the t-SNE components. The intermingling of red and blue dots suggests that the generated data overlaps significantly with the real data, indicating that the model successfully captures the overall distribution of the real data. Appendix B presents additional ablation studies.

\subsection{Use Case Scenario}
To evaluate the model's real-world applicability, we leveraged its autoregressive capability to generate standard 3.75\,s segments of DUS signals by sequentially inputting consecutive heartbeat FECG signals into the model. The generated DUS segments were first passed through a high-performing signal quality assessment model \cite{motie-shirazi_point--care_2024}. This model categorizes each 3.75\,s DUS signal into five distinct quality classes: good, poor, interference, talking, and silent. The quality assessment model classified all 303 segments generated by Auto-FEDUS as \textit{good quality}, indicating the model's ability to mimic the characteristics of the real data (Appendix B). Second, a fetal heart rate (FHR) estimation model \cite{rafiei_autofhr_2024} processed the generated signals and calculated the heart rate for each segment. Figure \ref{fig:PSD} (c) presents a Bland-Altman plot to analyze the agreement between the label FHR and those of the synthesized segments. The results, discussed in more detail in Appendix B, indicate a high level of agreement with minimal bias between the two sets of segments. 

\section{Conclusions and Future Works}
Analyzing DUS signals, which can be swiftly and easily captured using basic devices, provides unique advantages for various tasks in fetal health monitoring; however, the practical application of automated models using these signals faces significant obstacles due to the scarcity of high-quality labeled datasets and the labor-intensive process of data handling. In addition, the complex, broadband nature of raw DUS signals and their unpredictable variations make their synthetic generation challenging. Here, we develop Auto-FEDUS, a solution for mapping FECG signals into their corresponding DUS signal through an autoregressive scheme. Experimental results demonstrate a clear advantage of the developed model in effectively capturing the key characteristics of DUS signals. It generates high-quality, realistic DUS signals that closely align with actual data, specifically achieving similar quality classifications and FHR estimation outcomes. This advancement can significantly accelerate the development of effective and robust machine learning models for fetal health monitoring.

For future work, we will leverage the available FECG data for this cross-modal signal extrapolation to assess the benefits of generated DUS signals in developing and improving the performance of various fetal monitoring tasks, including gestational age estimation and heartbeat segmentation. Particularly, we aim to mitigate bias toward specific cohorts and conditions that are underrepresented in datasets and often do not include the target groups that most need enhanced monitoring. Furthermore, a key plan is to enhance Auto-FEDUS's capability to conditionally account for a range of factors, including demographic considerations, signal quality, and the potential presence of birth defects. Once additional cross-setting datasets become available, the model's performance can be further evaluated and streamlined for deployment on portable devices, enabling real-time generation.

\section{Acknowledgments}
This work was supported by a Google.org AI for the Global Goals Impact Challenge Award, the National Institutes of Health, the Fogarty International Center and the Eunice Kennedy Shriver National Institute of Child Health and Human Development, under grant \# 1R21HD084114 and 1R01HD110480. N. Katebi is funded by a PREHS-SEED award grant \# K12ESO33593. G. D. Clifford is also partially supported by the National Center for Advancing Translational Sciences of the National Institutes of Health under Award \# UL1TR002378. The content is solely the responsibility of the authors and does not necessarily represent the official views of the National Institutes of Health. 

\bibliography{aaai25}

\begin{thebibliography}{35}
\providecommand{\natexlab}[1]{#1}

\bibitem[{Abdelgaber et~al.(2023)Abdelgaber, Salah, Omer, Farghal, and Mubarak}]{abdelgaber_subject-independent_2023}
Abdelgaber, K.~M.; Salah, M.; Omer, O.~A.; Farghal, A. E.~A.; and Mubarak, A.~S. 2023.
\newblock Subject-independent per beat {PPG} to single-lead {ECG} mapping.
\newblock \emph{Information}, 14(7): 377.

\bibitem[{Agostinelli et~al.(2017)Agostinelli, Marcantoni, Moretti, Sbrollini, Fioretti, Di~Nardo, and Burattini}]{agostinelli2017noninvasive}
Agostinelli, A.; Marcantoni, I.; Moretti, E.; Sbrollini, A.; Fioretti, S.; Di~Nardo, F.; and Burattini, L. 2017.
\newblock Noninvasive fetal electrocardiography part I: {Pan-Tompkins}' algorithm adaptation to fetal {R-peak} identification.
\newblock \emph{The open biomedical engineering journal}, 11: 17.

\bibitem[{Andreotti et~al.(2016)Andreotti, Behar, Zaunseder, Oster, and Clifford}]{andreotti_open-source_2016}
Andreotti, F.; Behar, J.; Zaunseder, S.; Oster, J.; and Clifford, G.~D. 2016.
\newblock An open-source framework for stress-testing non-invasive foetal {ECG} extraction algorithms.
\newblock \emph{Physiological Measurement}, 37(5): 627.

\bibitem[{Arjovsky, Chintala, and Bottou(2017)}]{arjovsky_wasserstein_2017}
Arjovsky, M.; Chintala, S.; and Bottou, L. 2017.
\newblock {W}asserstein generative adversarial networks.
\newblock In \emph{Proceedings of the 34th International Conference on Machine Learning}, 214--223.

\bibitem[{Banerjee et~al.(2014)Banerjee, Sinha, Choudhury, and Visvanathan}]{banerjee2014photoecg}
Banerjee, R.; Sinha, A.; Choudhury, A.~D.; and Visvanathan, A. 2014.
\newblock {PhotoECG}: Photoplethysmographyto estimate {ECG} parameters.
\newblock In \emph{IEEE International Conference on Acoustics, Speech and Signal Processing (ICASSP)}, 4404--4408.

\bibitem[{Chen, Li, and Zhang(2023)}]{chen2023deep}
Chen, W.; Li, Z.; and Zhang, G. 2023.
\newblock A Deep Learning Method to Translate Phonocardiogram ({PCG}) Signals to Electrocardiogram ({ECG}) Waveforms.
\newblock In \emph{2023 IEEE 13th International Conference on CYBER Technology in Automation, Control, and Intelligent Systems (CYBER)}, 865--869.

\bibitem[{Garcia-Canadilla et~al.(2014)Garcia-Canadilla, Rudenick, Crispi, Cruz-Lemini, Palau, Camara, Gratacos, and Bijens}]{Garcia-Canadilla2014}
Garcia-Canadilla, P.; Rudenick, P.~A.; Crispi, F.; Cruz-Lemini, M.; Palau, G.; Camara, O.; Gratacos, E.; and Bijens, B.~H. 2014.
\newblock A Computational Model of the Fetal Circulation to Quantify Blood Redistribution in Intrauterine Growth Restriction.
\newblock \emph{PLoS Computational Biology}, 10(6): e1003667.

\bibitem[{Gibb and Arulkumaran(2023)}]{gibb_fetal_2023}
Gibb, D.; and Arulkumaran, S. 2023.
\newblock \emph{Fetal Monitoring in Practice}.
\newblock Elsevier, Churchill Livingstone, 5th edition.

\bibitem[{Harfiya, Chang, and Li(2021)}]{harfiya2021continuous}
Harfiya, L.~N.; Chang, C.-C.; and Li, Y.-H. 2021.
\newblock Continuous blood pressure estimation using exclusively photopletysmography by {LSTM}-based signal-to-signal translation.
\newblock \emph{Sensors}, 21(9): 2952.

\bibitem[{Ibtehaz et~al.(2022)Ibtehaz, Mahmud, Chowdhury, Khandakar, Salman~Khan, Ayari, Tahir, and Rahman}]{ibtehaz2022ppg2abp}
Ibtehaz, N.; Mahmud, S.; Chowdhury, M.~E.; Khandakar, A.; Salman~Khan, M.; Ayari, M.~A.; Tahir, A.~M.; and Rahman, M.~S. 2022.
\newblock {PPG2ABP}: Translating photoplethysmogram ({PPG}) signals to arterial blood pressure ({ABP}) waveforms.
\newblock \emph{Bioengineering}, 9(11): 692.

\bibitem[{Karras et~al.(2018)Karras, Aila, Laine, and Lehtinen}]{karras_progressive_2018}
Karras, T.; Aila, T.; Laine, S.; and Lehtinen, J. 2018.
\newblock Progressive growing of {GANs} for improved quality, stability, and variation.

\bibitem[{Katebi et~al.(2023)Katebi, Sameni, Rohloff, and Clifford}]{katebi_hierarchical_2023}
Katebi, N.; Sameni, R.; Rohloff, P.; and Clifford, G.~D. 2023.
\newblock Hierarchical attentive network for gestational age estimation in low-resource settings.
\newblock \emph{IEEE Journal of Biomedical and Health Informatics}, 27(5): 2501--2511.

\bibitem[{Kong et~al.(2024)Kong, Lee, Do, Park, Xu, Mao, and Chung}]{kong2024f}
Kong, N.~C.; Lee, D.; Do, H.; Park, D.~H.; Xu, C.; Mao, H.; and Chung, J. 2024.
\newblock {f-GAN}: A frequency-domain-constrained generative adversarial network for {PPG} to {ECG} synthesis.
\newblock \emph{arXiv preprint arXiv:2406.16896}.

\bibitem[{Marzbanrad et~al.(2017)Marzbanrad, Khandoker, Kimura, Palaniswami, and Clifford}]{10.3389/fphys.2017.00313}
Marzbanrad, F.; Khandoker, A.~H.; Kimura, Y.; Palaniswami, M.; and Clifford, G.~D. 2017.
\newblock Assessment of fetal development using cardiac valve intervals.
\newblock \emph{Frontiers in Physiology}, 8.

\bibitem[{Marzbanrad, Stroux, and Clifford(2018)}]{marzbanrad_cardiotocography_2018}
Marzbanrad, F.; Stroux, L.; and Clifford, G.~D. 2018.
\newblock Cardiotocography and beyond: a review of one-dimensional {Doppler} ultrasound application in fetal monitoring.
\newblock \emph{Physiological Measurement}, 39(8): 08TR01.

\bibitem[{Motie-Shirazi et~al.(2023)Motie-Shirazi, Sameni, Rohloff, Katebi, and Clifford}]{motie-shirazi_point--care_2024}
Motie-Shirazi, M.; Sameni, R.; Rohloff, P.; Katebi, N.; and Clifford, G.~D. 2023.
\newblock Point-of-care real-time signal quality for fetal {Doppler} ultrasound using a deep learning approach.
\newblock In \emph{Conference on Machine Learning for Health (ML4H)}.

\bibitem[{Oord et~al.(2016)Oord, Dieleman, Zen, Simonyan, Vinyals, Graves, Kalchbrenner, Senior, and Kavukcuoglu}]{oord2016wavenet}
Oord, A. v.~d.; Dieleman, S.; Zen, H.; Simonyan, K.; Vinyals, O.; Graves, A.; Kalchbrenner, N.; Senior, A.; and Kavukcuoglu, K. 2016.
\newblock {WaveNet}: A generative model for raw audio.
\newblock \emph{arXiv preprint arXiv:1609.03499}.

\bibitem[{Pennati, Bellotti, and Fumero(1997)}]{Pennati1997}
Pennati, M.; Bellotti, M.; and Fumero, R. 1997.
\newblock Mathematical modelling of the human foetal cardiovascular system based on {D}oppler ultrasound data.
\newblock \emph{Medical Engineering \& Physics}, 19(4): 327--335.

\bibitem[{Queyam, Pahuja, and Singh(2017)}]{Queyam2017}
Queyam, A.~B.; Pahuja, S.~K.; and Singh, D. 2017.
\newblock Simulation and analysis of umbilical blood flow using {M}arkov-based mathematical model.
\newblock \emph{International Journal of Intelligent Systems and Applications}, 9: 41.

\bibitem[{Rafiei et~al.(2024)Rafiei, Motie-Shirazi, Sameni, Clifford, and Katebi}]{rafiei_autofhr_2024}
Rafiei, A.; Motie-Shirazi, M.; Sameni, R.; Clifford, G.~D.; and Katebi, N. 2024.
\newblock AutoFHR: A Neural Temporal Model for Fetal Cardiac Activity Analysis.
\newblock In \emph{Conference on Machine Learning for Health (ML4H)}.

\bibitem[{Sameni(2012)}]{sameni2012oset}
Sameni, R. 2012.
\newblock {OSET}: the open-source electrophysiological toolbox.
\newblock \emph{Version 3.14 [Electronic Resource]}.

\bibitem[{Sameni et~al.(2007)Sameni, Clifford, Jutten, and Shamsollahi}]{sameni2007multichannel}
Sameni, R.; Clifford, G.~D.; Jutten, C.; and Shamsollahi, M.~B. 2007.
\newblock Multichannel {ECG} and noise modeling: Application to maternal and fetal {ECG} signals.
\newblock \emph{EURASIP Journal on Advances in Signal Processing}, 2007: 1--14.

\bibitem[{Sarkar and Etemad(2021)}]{sarkar_cardiogan_2021}
Sarkar, P.; and Etemad, A. 2021.
\newblock {CardioGAN}: attentive generative adversarial network with dual discriminators for synthesis of {ECG} from {PPG}.
\newblock \emph{Proceedings of the AAAI Conference on Artificial Intelligence}, 35(1): 488--496.

\bibitem[{Shome, Sarkar, and Etemad(2024)}]{shome2024region}
Shome, D.; Sarkar, P.; and Etemad, A. 2024.
\newblock Region-disentangled diffusion model for high-fidelity {PPG}-to-{ECG} translation.
\newblock In \emph{Proceedings of the AAAI Conference on Artificial Intelligence}, volume~38, 15009--15019.

\bibitem[{Stroux et~al.(2017)Stroux, Redman, Georgieva, Payne, and Clifford}]{stroux_dopplerbased_2017}
Stroux, L.; Redman, C.~W.; Georgieva, A.; Payne, S.~J.; and Clifford, G.~D. 2017.
\newblock Doppler‐based fetal heart rate analysis markers for the detection of early intrauterine growth restriction.
\newblock \emph{Acta Obstetricia et Gynecologica Scandinavica}, 96(11): 1322--1329.

\bibitem[{Sulas et~al.(2021)Sulas, Urru, Tumbarello, Raffo, Sameni, and Pani}]{Sulas2021}
Sulas, E.; Urru, M.; Tumbarello, R.; Raffo, L.; Sameni, R.; and Pani, D. 2021.
\newblock A non-invasive multimodal foetal {ECG–Doppler} dataset for antenatal cardiology research.
\newblock \emph{Scientific Data}, 8: 30.

\bibitem[{Tian et~al.(2020)Tian, Zhu, Li, and Wu}]{tian2020cross}
Tian, X.; Zhu, Q.; Li, Y.; and Wu, M. 2020.
\newblock Cross-domain joint dictionary learning for {ECG} reconstruction from {PPG}.
\newblock In \emph{IEEE International Conference on Acoustics, Speech and Signal Processing (ICASSP)}, 936--940.

\bibitem[{{UN Inter-agency Group for Child Mortality Estimation (UN IGME)}(2020)}]{UNIGME2020}
{UN Inter-agency Group for Child Mortality Estimation (UN IGME)}. 2020.
\newblock {A neglected tragedy: the global burden of stillbirths, Report}.

\bibitem[{{UN Inter-agency Group for Child Mortality Estimation (UNIGME)}(2021)}]{UNIGME2021}
{UN Inter-agency Group for Child Mortality Estimation (UNIGME)}. 2021.
\newblock {Levels and trends in child mortality, Report}.

\bibitem[{Valderrama et~al.(2019)Valderrama, Stroux, Katebi, Paljug, Hall-Clifford, Rohloff, Marzbanrad, and Clifford}]{valderrama_open_2019}
Valderrama, C.~E.; Stroux, L.; Katebi, N.; Paljug, E.; Hall-Clifford, R.; Rohloff, P.; Marzbanrad, F.; and Clifford, G.~D. 2019.
\newblock An open source autocorrelation-based method for fetal heart rate estimation from one-dimensional {Doppler} ultrasound.
\newblock \emph{Physiological Measurement}, 40(2): 025005.

\bibitem[{Vo, El-Khamy, and Choi(2023)}]{vo2023ppg}
Vo, K.; El-Khamy, M.; and Choi, Y. 2023.
\newblock {PPG} to {ECG} signal translation for continuous atrial fibrillation detection via attention-based deep state-space modeling.
\newblock \emph{arXiv preprint arXiv:2309.15375}.

\bibitem[{Vo et~al.(2021)Vo, Naeini, Naderi, Jilani, Rahmani, Dutt, and Cao}]{vo_p2e-wgan_2021}
Vo, K.; Naeini, E.~K.; Naderi, A.; Jilani, D.; Rahmani, A.~M.; Dutt, N.; and Cao, H. 2021.
\newblock {P2E}-{WGAN}: {ECG} waveform synthesis from {PPG} with conditional wasserstein generative adversarial networks.
\newblock In \emph{Proceedings of the 36th {Annual} {ACM} {Symposium} on {Applied} {Computing}}, 1030--1036. New York, NY, USA.

\bibitem[{Yamamoto, Hiromatsu, and Ohtsuki(2020)}]{yamamoto_ecg_2020}
Yamamoto, K.; Hiromatsu, R.; and Ohtsuki, T. 2020.
\newblock {ECG} signal reconstruction via {Doppler} sensor by hybrid deep learning model with {CNN} and {LSTM}.
\newblock \emph{IEEE Access}, 8: 130551--130560.

\bibitem[{Yu, Bouazizi, and Ohtsukil(2023)}]{yu_improving_2023}
Yu, D.; Bouazizi, M.; and Ohtsukil, T. 2023.
\newblock Improving heart rate range classification using {Doppler} radar with {GAN}-based data augmentation.
\newblock In \emph{{IEEE} {Global} {Communications} {Conference} ({GLOBECOM})}, 3885--3890.

\bibitem[{Zhu et~al.(2021)Zhu, Tian, Wong, and Wu}]{zhu2021learning}
Zhu, Q.; Tian, X.; Wong, C.-W.; and Wu, M. 2021.
\newblock Learning your heart actions from pulse: {ECG} waveform reconstruction from {PPG}.
\newblock \emph{IEEE Internet of Things Journal}, 8(23): 16734--16748.

\end{thebibliography}

\appendix

\section{Appendix A. Data Processing}\label{apd:pm}
\begin{figure}[htbp]
  \centering
  \includegraphics[width=0.5\textwidth]{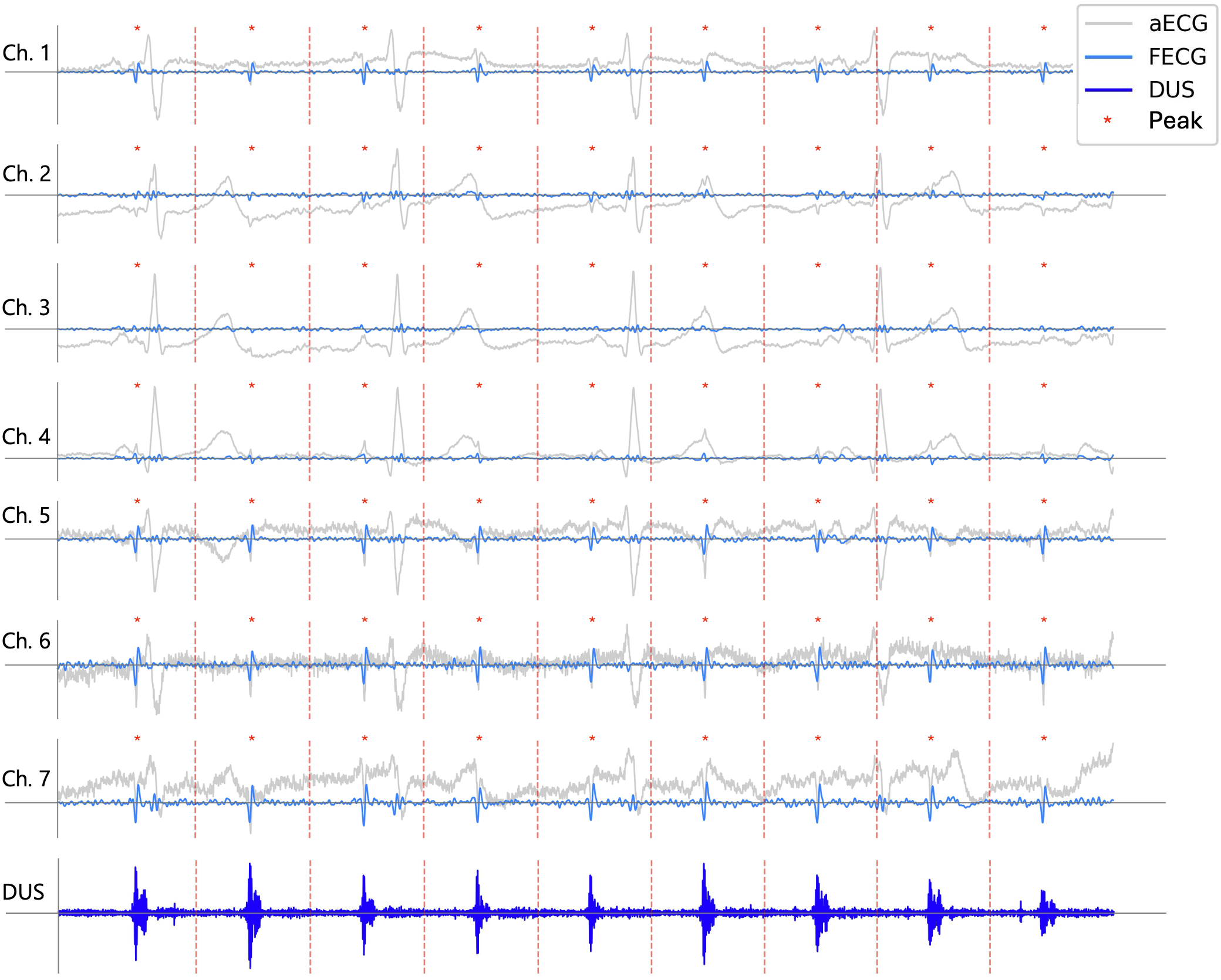}
  \caption{Different signals and the location of heartbeats for a random subject in the dataset. aECG: abdomen ECG, FECG: fetal ECG, DUS: Doppler Ultrasound, Peak: FECG peak, Ch.: Maternal ECG channel number}
  \label{fig:data}
\end{figure}

Figure \ref{fig:data} presents a 3.75\,s segment of physiological signals from a random subject, showcasing data from multiple modalities and highlighting the temporal correspondence between abdominal ECG, FECG, and DUS signals. The gray traces represent the abdominal ECG signals, while the overlaid blue traces highlight the extracted FECG signals across seven channels. The bottom panel displays the DUS signal, showing periodic waveforms that correspond to cardiac activity. Red asterisks mark the identified FECG R-peaks, and vertical dashed lines separate consecutive heartbeats.

\begin{figure}[h!]
  \centering
  \includegraphics[width=0.46\textwidth]{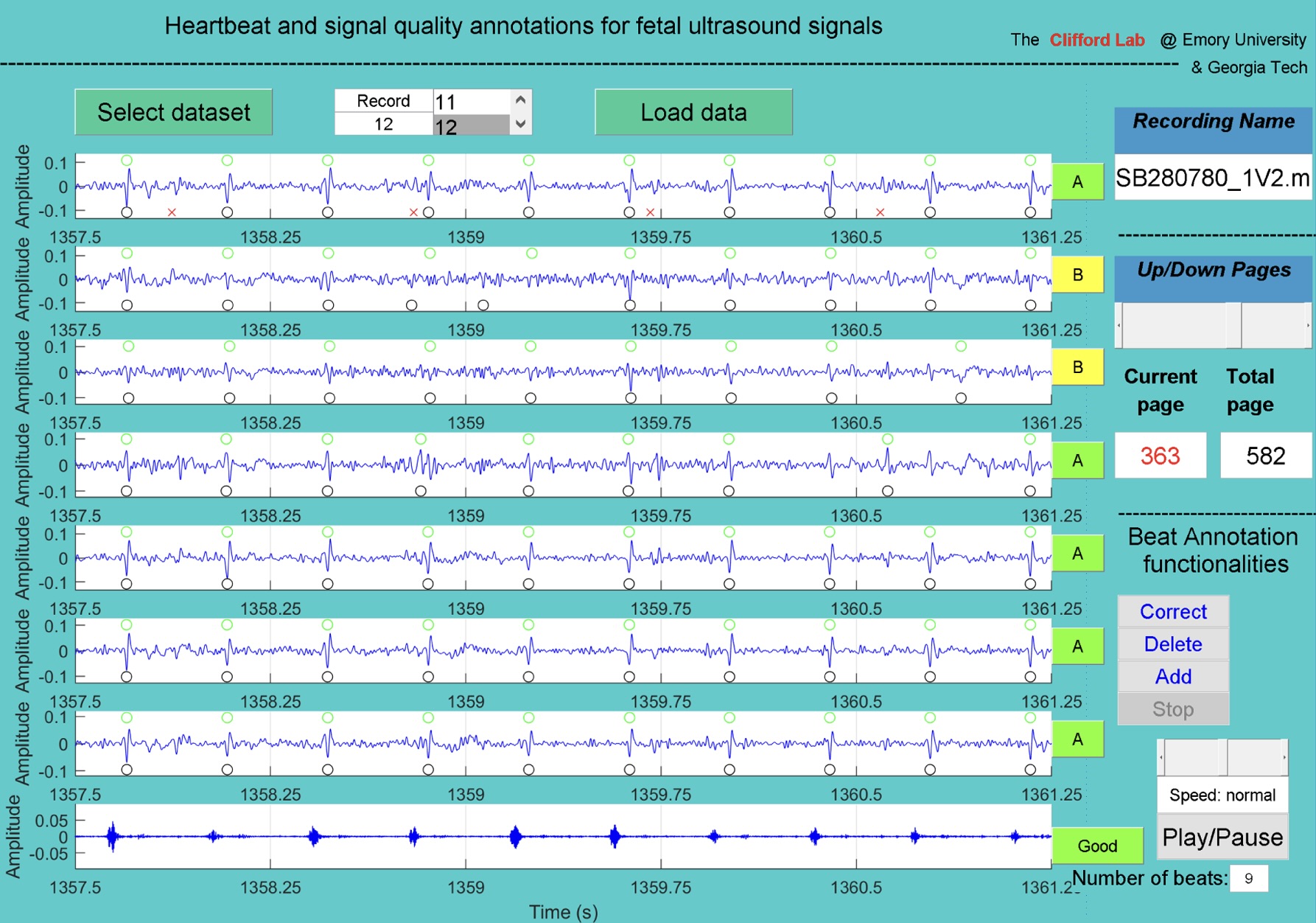}
  \caption{The custom MATLAB graphical user interface developed for signal quality assessment and peak detection.}
  \label{fig:gui}
\end{figure}

The developed MATLAB graphical user interface for the visualization and annotation of FECG and DUS signals is shown in Figure \ref{fig:gui}. This interface features multiple signal panels for the DUS and extracted seven-channel FECG signals. Users can navigate through different records using a dropdown menu and load the corresponding signals for detailed analysis. Key functionalities include classification labels for signal quality assessment, overlaid markers for heartbeat detection, and playback capabilities for reviewing signals. The detection of fetal peaks (black circles) and maternal beats (red crosses) can initially be performed automatically. Using the buttons on the right side of the interface, and leveraging the automatic detection as a preliminary guide, two independent annotators assigned SQIs to the signals and reviewed and adjusted peak placements (green circles) as needed.

\begin{figure*}[h!]
  \centering
  \includegraphics[width=0.98\textwidth]{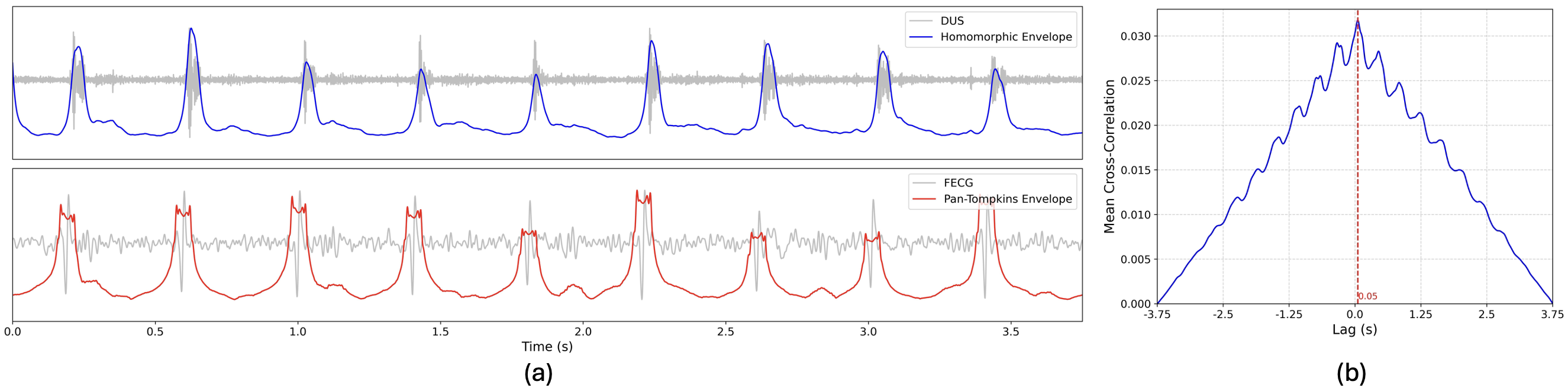}
  \caption{(a) DUS and FECG signals and their envelopes
  (b) Mean cross-correlation between the envelopes of DUS and FECG signals across all pairs and different time lags with highlighted strongest temporal alignment.}
  \label{fig:lag}
\end{figure*}

\begin{table*}[h!]
\centering
\caption{The results of the evaluation indicators for the Auto-FEDUS model in extrapolating longer signal segments based on leave-one-out cross-validation.}
\label{table:ab}
\fontsize{9}{11}\selectfont
\begin{tabular}{lcccccccc}
\toprule
\textbf{Model} & \textbf{RMSE} & \textbf{MAE} & \textbf{KLD} & \textbf{SE} & \textbf{PSDD} & \textbf{CD} & \textbf{SF} & \textbf{FD} \\ \midrule
Auto-FEDUS-2 beats & 0.19$\pm$0.01 & 0.13$\pm$0.01 & 0.04$\pm$0.01 & 0.34$\pm$0.35 & 0.19$\pm$0.11 & 85.98$\pm$37.03 & 0.02$\pm$0.01 & 6.19$\pm$0.36 \\ 

Auto-FEDUS-3 beats & 0.19$\pm$0.01 & 0.14$\pm$0.01 & 0.03$\pm$0.01 & 0.52$\pm$0.34 & 0.15$\pm$0.11 & 85.72$\pm$44.77 & 0.03$\pm$0.01 & 7.29$\pm$0.48 \\ 

\bottomrule
\end{tabular}
\label{tab:ablation}
\end{table*}

\begin{figure*}[t!]
  \centering
  \includegraphics[width=0.88\textwidth]{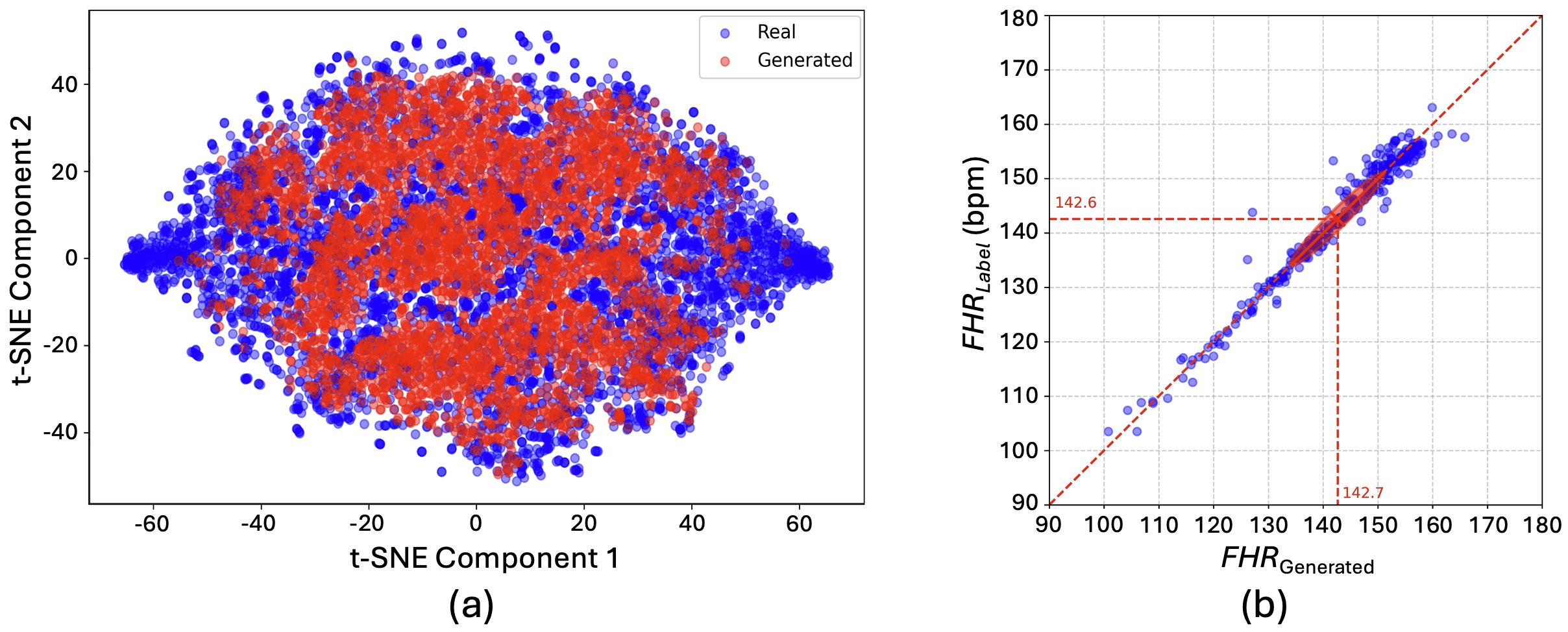}
  \caption{Reliability and comparative analysis. (a) t-SNE visualization (b) Scatter plot of the correlation between the real and generated FHR values.}
  \label{fig:ablation}
\end{figure*}

In simultaneous recordings of DUS and FECG, an inherent lag exists due to the precedence of electrical activity over mechanical activity, as well as differences in acquisition and processing mechanisms. To ensure that the lag remains consistent across all subjects and does not affect our analysis, we computed the homomorphic envelope for the DUS signals and the envelope derived from the integration step of the Pan-Tompkins QRS detection algorithm for the FECG signals \cite{agostinelli2017noninvasive}. We then estimated the temporal lag between these envelopes by calculating their cross-correlation and aligned each pair of signals accordingly. Figure \ref{fig:lag} (a) illustrates the signals along with their envelopes, while Figure \ref{fig:lag} (b) shows the mean cross-correlation between the envelopes across different time lags.

\section{Appendix B. Ablation Study and Practical Assessment}

We investigated the performance of the Auto-FEDUS model in generating longer segments of signals at each step. For this purpose, we extracted more consecutive heartbeats of FECG signals along with their corresponding simultaneous DUS waveforms and then performed subject-wise training and evaluation. The findings revealed that the model was able to extrapolate up to three beats at each step while maintaining acceptable signal morphology. However, when the extrapolation task involved more than three beats, the generated signals exhibited notable degradation in quality and fidelity. Table \ref{tab:ablation} summarizes the evaluation metrics for the 2- and 3-beat mapping scenarios. 

\begin{figure*}[h!]
  \centering
  \includegraphics[width=0.98\textwidth]{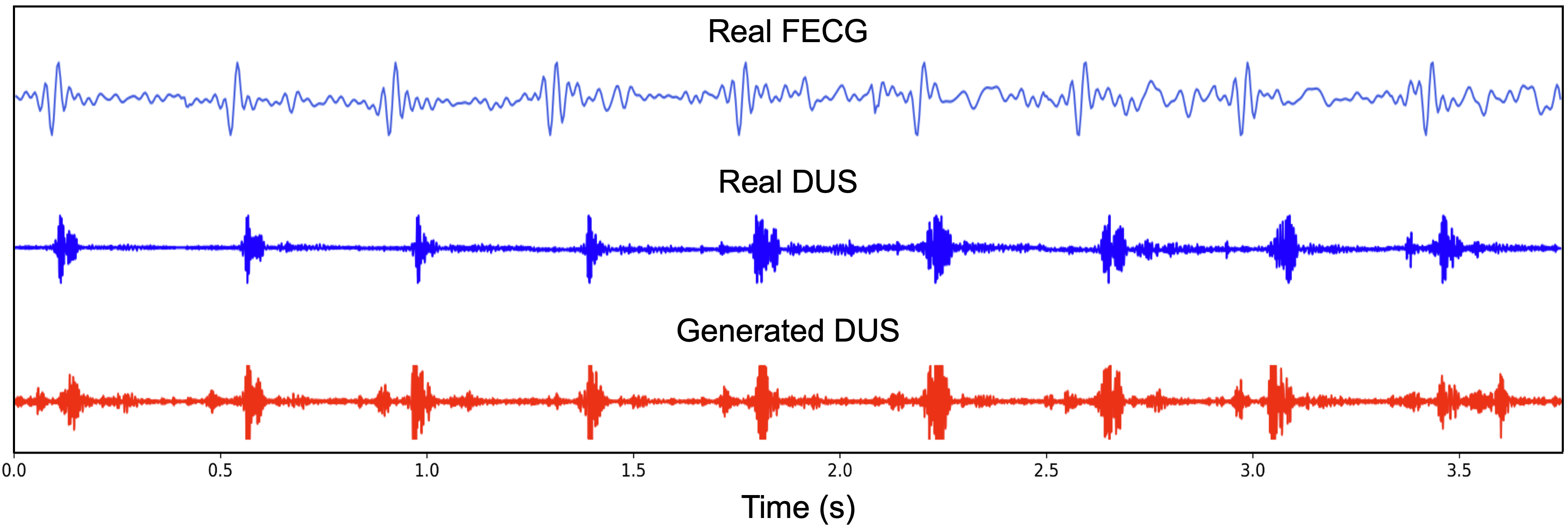}
  \caption{A 3.75 s segment of real FECG data and its corresponding real and generated DUS waveforms.}
  \label{fig:seg}
\end{figure*}

Moreover, we developed a CNN-based autoencoder designed for signal reconstruction. This model processed both real and generated DUS waveforms as input, aiming to accurately reconstruct these signals. The architecture of the encoder comprised three convolutional blocks, each consisting of a CNN layer followed by max pooling and batch normalization. Similarly, the decoder mirrored this structure with three transposed CNN layers, each followed by batch normalization. The objective of this experiment was to determine whether a high-performance reconstruction model with low-dimensional latent space representation could effectively distinguish between real and generated signals. Upon training, the model demonstrated a near-perfect capability to reconstruct the input signals. Nonetheless, visualization of the t-SNE plot revealed that the signals' representations are interspersed across the components without clear distinctions (Figure \ref{fig:ablation} (a)). This indicates that the real and generated signals share significantly similar characteristics, making them challenging to distinguish.

To further assess how closely the generated data replicate the morphology and characteristics of the real signals, we presented a scatter plot that illustrates the relationship between the estimated FHR values of the synthesized 3.75\,s segments and their corresponding real labels (Figure \ref{fig:ablation} (b)). The DUS-based FHR estimation model \cite{rafiei_autofhr_2024} provided estimated values for every generated segment, while the label heart rate was derived from real FECG R-peak locations. As shown, the generated data accurately captures the pseudo-periodic pattern of the signals, effectively learning to generate heartbeats with precise timing across both low and high heart rate scenarios. 
Additionally, Table \ref{tab:comparison} details a performance comparison of the FHR estimation model on its reported real test data and our generated signals \cite{rafiei_autofhr_2024}. The Bland-Altman measure, which represents the maximum difference between the model's estimations and the label FHR values, was calculated as the maximum of the mean difference plus or minus 1.96 times the standard deviation of the differences. Prediction interval coverage probability (PICP) was calculated as the percentage of data points where the difference between the model's estimation and label FHR fell within ±5 beats per minute (bpm) error range. The results indicate that the FHR estimation model performs similarly across different metrics for both real and generated data, further validating the realism of the synthesized signals. Figure \ref{fig:seg} shows a 3.75\,s segment of real FECG signal along with its corresponding real DUS signal and autoregressively generated DUS signal for a random subject, which is used as the standard duration for DUS-based fetal cardiac activity monitoring tasks.

\begin{table}[htbp]
\centering
\caption{Performance comparison of the FHR estimation model for the real and generated data.}
\fontsize{9}{11}\selectfont
\begin{tabular}{lcccc}
\toprule
\textbf{Data} & \textbf{Bland-Altman} & \textbf{RMSE} & \textbf{MAE} & \textbf{PICP} \\ \hline
Real & 4.5 bpm & 2.2 bpm & 1.1 bpm & 98.1\% \\
Generated & 4.5 bpm & 2.3 bpm & 1.5 bpm & 96.7\% \\
\bottomrule
\end{tabular}
\label{tab:comparison}
\end{table}

\end{document}